\def\year{2019}\relax
\documentclass[letterpaper]{article} 
\usepackage{aaai19}  
\usepackage{times}
\usepackage{helvet}
\usepackage{courier}
\usepackage{url}
\usepackage{graphicx}
\usepackage{amsmath}
\usepackage{amssymb}
\usepackage{multirow} 
\usepackage{appendix} 
\usepackage[usenames, dvipsnames]{color} 
\begin{document}

\title{Knowledge-driven Encode, Retrieve, Paraphrase for \\ Medical Image Report Generation}
\author{
Christy Y. Li\thanks{This work was done when Christy Y. Li was at Petuum, Inc.}\textsuperscript{\rm 1}, 
Xiaodan Liang\thanks{Corresponding author.}\textsuperscript{\rm 2},
Zhiting Hu\textsuperscript{\rm 2}, 
Eric P. Xing\textsuperscript{\rm 3}\\
\textsuperscript{\rm 1}Duke University,
\textsuperscript{\rm 2}Carnegie Mellon University ,
\textsuperscript{\rm 3}Petuum, Inc \\
yl558@duke.edu, \{xiaodan1,zhitingh\}@cs.cmu.edu, eric.xing@petuum.com. 
}

\maketitle
\begin{abstract}
Generating long and semantic-coherent reports to describe medical images poses great challenges towards bridging visual and linguistic modalities, incorporating medical domain knowledge, and generating realistic and accurate descriptions. We propose a novel \textit{Knowledge-driven Encode, Retrieve, Paraphrase} (KERP) approach which reconciles traditional knowledge- and retrieval-based methods with modern learning-based methods for accurate and robust medical report generation. 
Specifically, KERP decomposes medical report generation into explicit medical abnormality graph learning and subsequent natural language modeling. 
KERP first employs an \textit{Encode} module that transforms visual features into a structured abnormality graph by incorporating prior medical knowledge; then a \textit{Retrieve} module that retrieves text templates based on the detected abnormalities; and lastly, a \textit{Paraphrase} module that rewrites the templates according to specific cases.
The core of KERP is a proposed generic implementation unit---Graph Transformer (GTR) that dynamically transforms high-level semantics between graph-structured data of multiple domains such as knowledge graphs, images and sequences. 
Experiments show that the proposed approach generates structured and robust reports supported with accurate abnormality description and explainable attentive regions, achieving the state-of-the-art results on two medical report benchmarks, with the best medical abnormality and disease classification accuracy and improved human evaluation performance. 
\end{abstract}

\section{Introduction}

Beyond the traditional image captioning task~\cite{xu2015show,karpathy2015deep,rennie2016self} that produces single-sentence descriptions, generating long and semantic-coherent stories or reports to describe visual contents (e.g., images, videos) has recently attracted increasing research interests~\cite{liang2017recurrent,huang2016visual,krause2017hierarchical}, 
and is posed as a more challenging and realistic goal towards bridging visual patterns with human linguistic descriptions. 
Particularly, an outstanding challenge in modeling long narrative from visual content is to balance between knowledge discovery and language modeling~\cite{karpathy2015deep}.
Current visual text generation approaches tend to generate plausible sentences that look natural by the language model but poor at finding visual groundings. 
Although some approaches have been proposed to alleviate this problem~\cite{neural-baby-talk,anderson2017bottom,liang2017recurrent}, most of them ignore the internal knowledge structure of the task at hand.
However, most real-world data and problems exhibit complex and dynamic structures such as intrinsic relations among discrete entities under nature's law~\cite{taskar2004max,hu2016harnessing,strubell2018linguistically}. 
Knowledge graph, as one of the most powerful representations of dynamic graph-structured knowledge~\cite{mitchell2018never,bizer2011linked}, complements the learning-based approaches by explicitly modeling the domain-specific knowledge structure and relational inductive bias.  
Knowledge graph also allows incorporating priors, which is proven useful for tasks where universal knowledge is desired or certain constraints have to be met~\cite{battaglia2018relational,liang2018symbolic,hu2018deep,wang2018zero-shot}. 
 
\begin{figure}
\centering 
\includegraphics[scale=0.9,trim={0 11.2cm 2cm 0},clip]{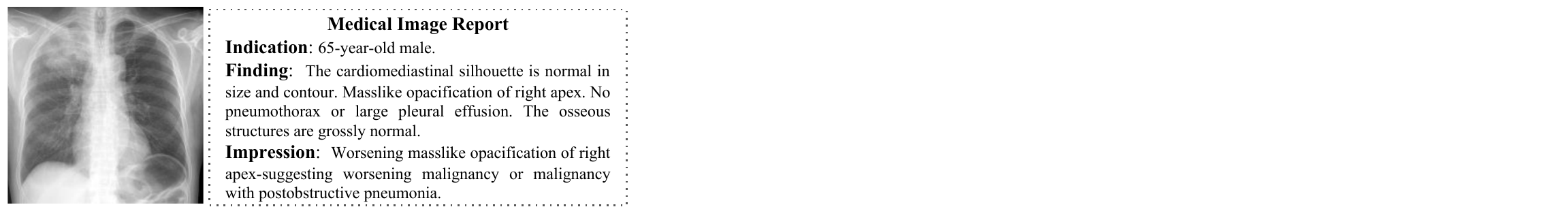}
\caption{Example of medical image report} 
\label{fig:xrayimage}
\end{figure}

As an emerging task of long text generation of practical use, \emph{medical image report generation}~\cite{li2018hybrid,jing2017automatic} must satisfy more critical protocols and ensure the correctness of medical terminology usage. 
As shown in Figure~\ref{fig:xrayimage}, 
a medical report consists of a finding section describing medical observations in details of both normal and abnormal features, an impression or conclusion sentence indicating the most prominent medical observation, and peripheral sections such as patient’s information and indications. Among these sections, the finding section is considered as the most important component and is expected to
1) cover contents of key relevant aspects such as heart size, lung opacity, and bone structure; 
2) correctly detect any abnormalities and support with details such as the location and shape of the abnormality; 
3) describe potential diseases such as effusion, pneumothorax and consolidation.  

It is often observed that, to write a medical image report, radiologists first check a patient's images for abnormal findings, then write reports by following certain patterns and templates, and adjusting statements in the templates for each individual case when necessary~\cite{hong2013content}. To mimic this procedure, we propose to formulate medical report writing as a knowledge-driven encode, retrieve, paraphrase (KERP) process. In particular, KERP first invokes an \textit{Encode} module to transform visual features of medical images into an abnormality graph where each node represents a possible clinical abnormality designed by prior medical knowledge, and the features of which depict semantics of the abnormality that can be observed from the visual input (e.g., normal or abnormal, size, location). The correlation of abnormality nodes is further encoded as edge weights of the abnormality graph so that the relations among different abnormal findings are considered when making a clinical diagnostic decision.  Then KERP retrieves a sequence of templates according to the detected abnormalities via a \textit{Retrieve} module. The words of the retrieved templates are further expanded and paraphrased into a report by a \textit{Paraphrase} module which enriches the templates with details and corrects false information if any. 
 
As most real-world data (e.g., images, sequence of text tokens, knowledge graphs, convolutional feature maps) can be represented as graphs, we further propose a novel and generic implementation unit---Graph Transformer (GTR) which dynamically transforms among multi-domain graph-structured data. We further equip GTR with attention mechanism for learning robust graph structure, as well as incorporating prior knowledge structure. By invoking GTR, KERP can transform robustly from visual features to an abnormality graph (with the \textit{Encode} module), then to sequences of templates (with the \textit{Retrieve} module), and lastly to sequences of words (with the \textit{Paraphrase} module).
 
We conduct extensive experiments on two medical image report dataset~\cite{iuxray}. Our KERP achieves the state-of-the-art performance on both datasets under both automatic evaluation metrics and human evaluation. KERP also achieves best performance on abnormality classification. Experiments show that KERP not only generates structured and robust reports supported with accurate abnormality prediction, but also produces explainable attentive regions which is crucial for interpretative diagnosis. 

\section{Related Work}

\subsubsection{Medical report generation.} 
Machine learning for healthcare has been widely recognized in both academia and industry as an area of high impact and potential. Automatic generation of medical image reports, as one of the key applications in the field, is gaining increasing research interest~\cite{li2018hybrid,wang2018tienet,jing2017automatic}. The task differs from other tasks such as summarization where summaries tend to be more diverse without clear templates or internal knowledge structure; and image captioning where usually a single sentence is desired. 

\subsubsection{Graph neural networks.}
Graph neural networks (GNN) have gained increasing research interests~\cite{defferrard2016convolutional,kipf2016semi,monti2017geometric}.   
However, most existing methods learn to encode the input feature into higher-level feature through selective attention over the object itself~\cite{wang2017non,parmar2018image,velickovic2017graph}, while our method works on multiple graphs, and models not only the data structure within the same graph but also the transformation rules among different graphs.

\subsubsection{Hybrid retrieval-generation approach.}
Combining traditional retrieval-based and modern generation-based methods for (long) text generation~\cite{li2018hybrid,guu2017generating,retrieve-rerank-rewrite,hu2017toward} has gained increasing research interests. 
Our work differs from previous work in that: 1) we develop an encoding procedure that explicitly learns the graph structure of medical abnormalities; 2) the \textit{retrieve} and \textit{rerank} is formulated as one joint, comprehensive process and implemented via a novel and generic unit--Graph Transformer.

\section{Graph Transformer (GTR)} 
\label{sec:3}

We start by describing Graph Transformer (GTR) which transforms a graph into another graph for encoding features into higher-level semantics within the same graph type, or translating features of one graph (e.g., knowledge graph) into another one (e.g., sequence of words).
First, we represent a graph as $G = (V, E)$. Here $V = \{\textbf{v}_i\}_{i=1:N}$ is a set of nodes where each $\textbf{v}_i\in \mathbb{R}^d$ represents a node's feature of dimension $d$, and $N$ is the number of nodes in the graph. $E = \{e_{i,j}\}_{i,j=[1,N]}$ is a set of edges between any possible pair of nodes. Here we study the setting where each edge is associated with a scalar value indicating closeness of nodes, while it is straightforward to extend the formalism to other cases where edges are associated with non-scalar values such as vectors.  

GTR takes a graph $G = (V, E)$ as input, and outputs another graph $G' = (V', E')$. 
Note that $G$ and $G'$ are two different graphs and can have different structures and characteristics (e.g., $N \neq N'$, $d \neq d'$, and $e_{i,j} \neq e'_{i,j}$). 
This differs from many previous methods~\cite{defferrard2016convolutional,kipf2016semi,velickovic2017graph} which are restricted to the same graph structures. 
For both source and target graph, the set of nodes $V$ and $V'$ has to be given in prior (e.g., the vocabulary size if the considered graph is sequences, abnormality nodes if the considered graph is an abnormality graph). 
We consider two scenarios for the edges among graph nodes: 1) the edges are provided in prior, and denoted as $e_{s_i, t_j}$ where $s_i$ is the $i_{th}$ node of source graph and $t_j$ is the $j_{th}$ node of target graph; 2) the edges are not provided, and thus source and target nodes are represented as fully connected with uniform weights. We assume $e_{s_i, t_j}$ as normalized, to avoid notation of averaging.

There are two types of message passing in GTR: from source graph to target graph (inter-graph message passing), and message passing within the same graph (intra-graph message passing). 

\subsubsection{Inter-graph message passing} 
To learn the source graph's knowledge, the features of source nodes are transformed and passed to target nodes with their corresponding edge weights. The formulation can be written as:  
\begin{small}
\begin{align}  
&\textbf{v}'_{j} = \textbf{v}'_{j} + \sigma(\sum\nolimits_{i=1}^N e_{s_i, t_j} \textbf{W}_s \textbf{v}_i) 
\label{eq:1}
\end{align}
\end{small}
where $\sigma$ is a nonlinear activation, and $\textbf{W}_s$ is a projection matrix of size $d'\times d$. 

Considering that the edge information between source and target graphs may not be available in many cases (e.g., translating a sequence of words into another sequence of words), we propose to learn edge weights automatically by an attention mechanism~\cite{vaswani2017attention}. In this way, target node update is enabled to consider the varying importance of the source nodes. Specifically,
\begin{small}
\begin{align} 
\small
& \hat{e}_{s_i, t_j} = \textit{Attention} (\mathbf{W}^a_s\mathbf{v}_i, \mathbf{W}^a_t\mathbf{v}'_j)
\end{align}   
\end{small}
where $\hat{e}_{s_i, t_j}$ is the attention weight of edge from source node $i$ to target node $j$; $\textbf{W}^a_s$ and $\textbf{W}^a_t$ are weights in attention mechanism to project nodes features of source graph and target graph to a common space of dimension $q$ respectively; and \textit{Attention}: $\mathbb{R}^q \xrightarrow{} \mathbb{R}$ is the attention mechanism that transforms the two projected features $\mathbf{W}^a_s\mathbf{v}_i, \mathbf{W}^a_t\mathbf{v}'_j \in R^q$ to a scalar $\hat{e}_{s_i, t_j}$ as the edge's attention weight. In our experiments, \textit{Attention} is parameterized as a scaled dot-product operation with multi-head attention~\cite{vaswani2017attention}.  

The attention weights are normalized over all source nodes for each target node, denoting the relative importance of each source node to a target node among all source nodes. The formulation can be written as: 
\begin{small}
\begin{align} 
\small
& \hat{e}_{s_i, t_j} = softmax_{s_i} (\hat{e}_{s_i, t_j}) = \frac{\exp{(\hat{e}_{s_i, t_j})}}{\sum_{k=1}^N \exp{(\hat{e}_{s_k, t_j})}} 
\end{align}
\end{small}
Once obtained, the normalized attention coefficients are be combined with prior edge weights to pass features of connected source nodes to target nodes. The combined features are served as the target node's updated features with source graph knowledge encoded. We adopt weighted sum of the learned attention edge weights and prior edge weights as final edge weights. Other methods such as multiplication of learned and prior edge weights followed by softmax also works. However, in our experiments, we observed that the first method performs better and avoids under-fitting. The formulation can be written as: 
\begin{small}
\begin{align} 
&\Tilde{e}_{s_i, t_j} = \lambda e_{s_i, t_j} + (1-\lambda) \hat{e}_{s_i, t_j}
\label{eq:5}
\end{align}
\end{small}
where $\lambda$ is a user-defined weight controlling importance of prior edges and learned edges. If $\lambda$ is set to 1, the edges between source graph and target graph are fixed, and no attention machanism is required. The formulation is then the same as Equation~\ref{eq:1}. If $\lambda$ is set to $0$, the edges between source graph and target graph are completely learned by the model. 
With the updated weight, one can obtain updated target nodes features via Equation~\ref{eq:1}.

\subsubsection{Intra-graph message passing}
Intra-graph message passing aims at modeling the correlation among nodes of the same graph, and fusing features according to the closeness between them. 
Specifically, a target node is updated by combining features of neighboring nodes and itself. The formulation can be written as: 
\begin{small}
\begin{align}  
&\textbf{v}'_{j} = \textbf{v}'_{j} + \sigma(\sum\nolimits_{i=1}^{N'} \Tilde{e}_{i,j} \textbf{W}_t \textbf{v}'_i) 
\label{eq:6}
\end{align}
\end{small}
where $\textbf{W}_t$ is weight to project features of target nodes from dimension $d$ to output dimension. 
To learn the edge weights through attention mechanism, one can directly apply Equations~\ref{eq:1}-\ref{eq:5} by changing source and target nodes notation to be of the same graph.

\subsubsection{GTR as a module} 
\label{gtrasmodule}

\begin{figure} 
\centering 
\includegraphics[scale=0.6,trim={0 9cm 15cm 0},clip]{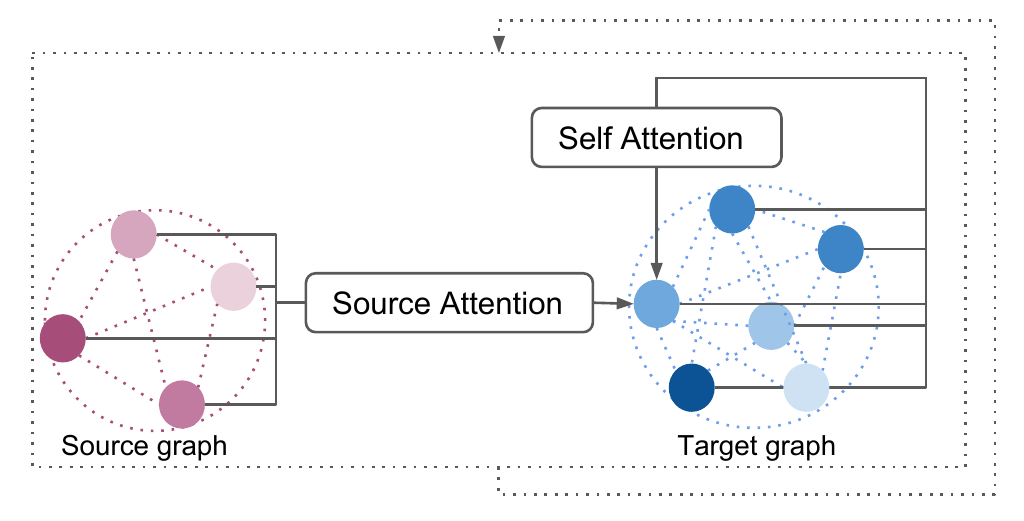}
   \caption{Architecture of \textit{Graph Transformer}. GTR evolves a target graph by recurrently performing \textit{Source Attention} on a source graph and \textit{Self Attention} on itself. The darkness of color of each graph node indicates the degree of attention the target node pays to. } 
\label{fig:gtr} 
\end{figure}

As shown in Figure~\ref{fig:gtr}, we formulate GTR as a module denoted as \textit{GTR} by first concatenating intra-graph message passing and inter-graph message passing into one step (that is, first conduct message passing within target graph, then conducting message passing from one or multiple source graphs), then stacking multiple such steps into one module in order to progressively convert target graph features into high-level semantics.  

\begin{figure*}[ht]
\centering 
\includegraphics[scale=0.7,trim={0 7.6cm 0 0},clip]{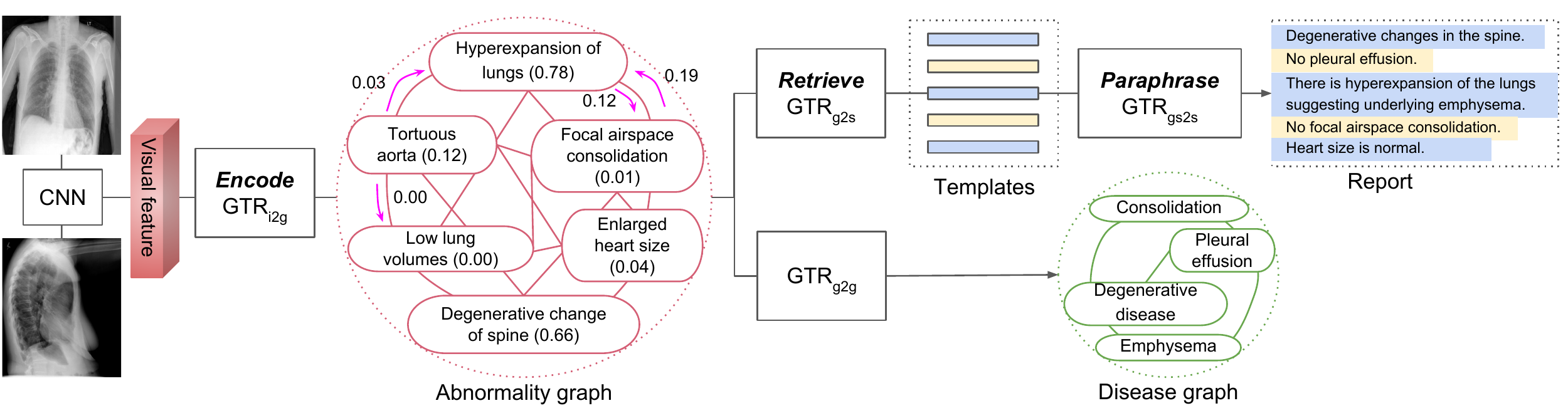}
   \caption{Architecture of KERP. Image features are first extracted from a CNN, and further encoded as an abnormality graph via \textit{Encode} $GTR_{i2g}$. \textit{Retrieve} $GTR_{g2s}$ decodes the abnormality graph as a template sequence, the words of which are then retrieved and paraphrased by \textit{Paraphrase} $GTR_{gs2s}$ as the generated report. Simultaneously, a $GTR_{g2g}$ decodes the abnormality graph as a disease graph, and predicts disease categories via extra classification layers. In the abnormality graph, values inside parentheses are probabilities of the corresponding nodes predicted by extra classification layers taking latent semantic features of nodes as input. Values along the directed arrows indicate attention scores of source nodes on target nodes. }
\label{fig:kerp}
\end{figure*}

\subsubsection{GTR for multiple domains} 

Most real-world data types (e.g., images, sequences, graphs) can be formulated as graph-structured. 
For example, a 2-dimentional image can be formulated as a graph whose nodes are pixels of the image where every node is connected with its neighboring pixel nodes; and a sequence of words can be formulated as a graph whose nodes are the individual words where edges among nodes are the consecutive relation among words. If global context of the data is considered, which is commonly adopted in attention mechanism~\cite{vaswani2017attention}, the graph nodes are then fully-connected. 
In the following, we describe the variants of \textit{GTR} for different data domains by first formulating data as graph-structured, and then perform GTR operations on it.  
In particular, we define $GTR_{i2g}$ as the variant of \textit{GTR} for transforming image features into graph's features; $GTR_{g2g}$ the variant of \textit{GTR} for transforming from a graph to another graph; $GTR_{g2s}$ the variant of \textit{GTR} for graph input and sequence output; and $GTR_{gs2s}$ the variant of GTR for graph and sequence input and sequence output. 
We use the variants of \textit{GTR} as building blocks of KERP for medical report generation, which is described in section\ref{sec:kerp}. 

\textbf{GTR for sequential input/output.}
To apply GTR for sequential input or output (e.g., a sequence of words, a sequence of retrieved items), we employ \textit{positional encoding}~\cite{vaswani2017attention} to GTR so as to add relative and absolute position information to the input or output sequence. Specifically, we use sine and cosine functions of different frequencies: 
\begin{align} 
&PE_{pos, 2i} = sin(pos/10000^{2i/d}) \\
&PE_{pos, 2i+1} = cos(pos/10000^{2i/d})
\end{align}
where $pos$ is the position and $i$ is the dimension. 
If both input and output are sequences, GTR is close to a \textit{Transformer}~\cite{vaswani2017attention} with prior edge weights.

\textbf{GTR for image input.}
We denote visual features of an image as $I \in R^{D, W, H}$ where D is the dimension of latent features, W and H is width and height. 
To apply GTR for image input, we first reshape visual features by flattening the 2-dimension into 1-dimension $R^{W \times H, D}$. Then each pixel is treated as graph node whose features are used as source graph features.

\textbf{GTR for multiple input graphs.}
For the cases where a target graph wants to learn from more than one source graphs, we extend \textit{GTR} to take into account multiple input source by replacing the single intre-graph message passing in each stacked layer of \textit{GTR} into multiple concatenated intre-graph message passing. 


\section{Knowledge-driven Encode, Retrieve, Paraphrase (KERP)} 
\label{sec:kerp}

It is observed that, to write a medical image report, radiologists first check a patient's images for abnormal findings, then write reports by following certain patterns and templates, and adjusting statements in the templates for each individual case when necessary~\cite{hong2013content}. To mimic this procedure, we propose to formulate medical report writing as a process of encoding, retrieval and paraphrasing. 
In particular, we first compile an off-the-shelf abnormality graph that contains large range of abnormal findings. We consider frequent abnormalities stem from thoracic organs as nodes in the abnormality graph. For example, "disappearance of costophrenic angle", "low lung volumes", and "blunted costophrenic angle". These abnormalities compose a major part of medical reports, and their detection quality would greatly impact the accuracy of the generated reports. Please see Appendix for detailed definition and examples. 
We also compile a template database that consists of a set of frequent sentences that cover descriptions of different abnormalities in the abnormality graph. For example, "there is hyperexpansion of lungs and flattening of the diaphragm consistent with copd" is a template for describing "hyperexpansion of lungs", "flattening of diaphragm" and "copd". 

Then we design separate modules for the purpose of encoding visual features as an abnormality graph, retrieving templates based on the detected abnormalities, and rewriting templates according to case-specific scenario. 
As described in Figure~\ref{fig:kerp}, a set of images are first fed into a CNN for extracting visual features which are then transformed into an abnormality graph via \textit{Encode} $GTR_{i2g}$. \textit{Retrieve} $GTR_{g2s}$ decodes the abnormality graph as a template sequence, the words of which are then retrieved and paraphrased by \textit{Paraphrase} $GTR_{gs2s}$ as the generated report. 
 
In addition, we design a disease graph containing common \emph{thorax} diseases (e.g., nodule, pneumonia and emphysema) which are commonly concluded from single or combined condition of abnormalities. For example, 
atelectasis may be concluded if "interval development of bandlike opacity in the left lung base" is present; 
consolidation and atelectasis may exist if there is "streaky and patchy bibasilar opacities", and "triangular density projected over the heart" without "typical findings of pleural effusion or pulmonary edema". In parallel to generating reports in the proposed model architecture, a $GTR_{g2g}$ is employed to transform the abnormality graph to a disease graph in order to predict common thorax diseases (right lower column of Figure~\ref{fig:kerp}) which can be useful as concluding information for medical reports.

\subsection{Encode: visual feature to knowledge graph} 

The \textit{Encode} module aims at encoding visual features as an abnormality graph. 
Assume an input image is encoded with a deep neural network into feature $\textbf{X} \in R^{WH, d_X}$ where $W$, $H$ and $d_X$ are width, height, and feature dimension, respectively. 
An abnormality graph is represented as a set of nodes of size $N$ with initialized features.
The latent features of each node can be used to predict occurrence of the abnormality via an additional classification layer.
By applying the variant of GTR for image input and graph output, denoted as $\textit{GTR}_{i2g}$, the updated node features can be written as: 
\begin{small}
\begin{align}  
&\textbf{h}_u = \textit{GTR}_{i2g}(\textbf{X}) \\
&\textbf{u} = \textit{sigmoid} (\textbf{W}_u \textbf{h}_u)
\end{align}
\end{small}
where $\textit{GTR}_{i2g}$ is the formulation of the variant of \textit{GTR} for image input and graph output described on page \pageref{gtrasmodule}, and $\textbf{W}_u$ is linear projection to transform latent feature $u$ into 1-d probability. $\textbf{h}_u = (\textbf{h}_{u_1}; \textbf{h}_{u_2}; ..., \textbf{h}_{u_N}) \in R^{N, d}$ is the set of latent features of nodes where $d$ is feature dimension. $\textbf{u} = (u_1, u_2, ..., u_N), y_i \in \{0, 1\}, i \in \{1,...,N\}$ denotes binary label for abnormality nodes. 

\subsection{Retrieve: knowledge graph to template sequence} 
Once the knowledge graph of abnormality attributes is obtained, it can be used to guide the retrieval process to harvest templates. A sequence of templates is represented as $\textbf{t} = (t_1, t_2, ..., t_{N_s})$ where $N_s$ is the maximum length of template sequence (also maximum number of sentences) and $t_i$ is index of templates where $i\in \{1,...,N_s\}$. 
By applying the variant of \textit{GTR} for graph input and sequential output, denoted as $\textit{GTR}_{g2s}$, the retrieved template sequence can be obtained as follows:
\begin{small}
\begin{align} 
&\textbf{h}_t = \textit{GTR}_{g2s}(\textbf{h}_u) \\
&\textbf{t} = \arg\max \textit{Softmax}(\textbf{W}_t \textbf{h}_t)
\end{align}
\end{small}
where $\textit{GTR}_{g2s}$ denotes the formulation of the variant of \textit{GTR} for graph input and sequence output, $\textbf{W}_t$ is linear projection to transform latent feature to template embedding, and $\arg\max \textit{Softmax}$ is the operation that selects index of maximum value after \textit{Softmax} function.

\subsection{Paraphrase: template sequence to report} 
\textit{Paraphrase} serves for two purposes: 1) refine templates with enriched details and possibly new case-specific findings; 2) convert templates into more natural and dynamic expressions. The first purpose is achieved by modifying information in the templates that is not accurate for specific cases, and the second purpose is achieved by robust language modeling for the same content. These two goals supplement each other in order to generate accurate and robust reports. 
 
Given the retrieved sequence of templates $\textbf{t}$ = $(t_1$, $t_2$, ..., $t_{N_s})$, the rewriting step generates a report consisting of a sequence of sentences $\textbf{R}$ = ($\textbf{r}_1$, $\textbf{r}_2$, ..., $\textbf{r}_{N_s}$) by subsequently attending to template words and the encoded knowledge graph (described on page 4). Each sentence $\textbf{r}_i$ = $(w_{i1}$, $w_{i2}$, ..., $w_{iN_w}$), $i\in \{1,..,N_s\}$ consists of a sequence of $N_w$ words. 
\begin{small}
\begin{align} 
&\textbf{h}_w = \textit{GTR}_{gs2s}(\textbf{h}_u, \textbf{t}) \\
&\textbf{R} = \arg\max\textit{Softmax} (\textbf{W}_w f(\textbf{h}_w))
\end{align}
\end{small}
where $\textit{GTR}_{gs2s}$ denotes the formulation of the variant of \textit{GTR} for graph and sequence input, and sequence output, $f$ denotes the operation of reshaping $\textbf{h}_w$ from $R^{N_s, N_w, d}$ to $R^{N_s*N_w, d}$, $\textbf{W}_w$ is linear projection to transform latent feature into word embedding, and $\arg\max\textit{Softmax}$ selects index of maximum value after $\textit{Softmax}$ function. 

\subsection{Disease classification} 

Abnormality attributes are frequently used for diagnosing diseases. Let $\textbf{z} = (z_1, z_2, ..., z_L)$ be a one-hot representation of disease labels where $L$ is the total number of disease labels, and $z_i \in \{0,1\}, i \in \{1, ...,L\}$. The multi-label disease classification can be written as: 
\begin{small}
\begin{align}  
&\textbf{h}_z = \textit{GTR}_{g2g}(\textbf{h}_u) \\
&\textbf{z} = \textit{sigmoid} (\textbf{W}_z \textbf{h}_z) 
\end{align} 
\end{small}
where $\textit{GTR}_{g2g}$ denotes the formulation of the variant of \textit{GTR} for graph input and graph output, $\textbf{W}_z$ is linear projection to transform disease nodes feature into 1-d probability. 

\subsection{Learning}

During paraphrasing, the retrieved templates \textbf{t}, instead of latent feature $\textbf{h}_t$, is used for rewriting. Sampling the templates of maximum predicted probability breaks the connectivity of differentiable back-propagation of the whole \textit{encode-retrieve-paraphrase} pipeline. To alleviate this issue, we first train the \textit{Paraphrase} with ground truth templates, and then with sampled templates generated by \textit{Retrieval} module. This minimizes the negative effect of dis-connectivity and make better test-time performance by letting the model accommodate to its own generated templates. Besides, given that templates serve as starting points instead of ground truth for rewriting, the prediction of templates does not have to be highly accurate as the \textit{Paraphrase} module needs to learn to paraphrase any suitable templates.

\begin{table*}[ht] 
\centering   
\small
\begin{tabular}{l|l|ccccccc}
\hline  
Dataset & Model & CIDEr & ROUGE-L  & BLEU-1 & BLEU-2 & BLEU-3 & BLEU-4  & Hit (\%)  \\ \hline 
\multirow{8}{*}{\scriptsize{\textbf{IU X-Ray}}} 
& CNN-RNN & 0.294 & 0.307 & 0.216 & 0.124 & 0.087 & 0.066 & -- \\ 
&LRCN &  0.285 & 0.307 & 0.223 & 0.128 & 0.089 & 0.068 & -- \\ 
&AdaAtt &  0.296 & 0.308 & 0.220 & 0.127 & 0.089 & 0.069 & -- \\ 
&Att2in &  0.297 & 0.307 & 0.224 & 0.129 & 0.089 & 0.068  & -- \\ 
& CoAtt* & 0.277 & \textbf{0.369} & 0.455 & 0.288 & 0.205 & 0.154 & 24.100\\
& HRGR-Agent & \textbf{0.343} & 0.322 & 0.438 & 0.298 & 0.208 & 0.151 \\
\cline{2-9}   
& KER & 0.318 & 0.335 & 0.455 & 0.304 & 0.210 & -- \\ 
& KERP & 0.280 & 0.339 & \textbf{0.482} & \textbf{0.325} & \textbf{0.226} & \textbf{0.162} & \textbf{57.425} \\ 
\hline 
\hline
\multirow{7}{*}{\scriptsize{\textbf{CX-CHR}}}  
& CNN-RNN & 1.580 & 0.578 & 0.592 & 0.506 & 0.450 & 0.411 & -- \\ 
&LRCN & 1.589 & 0.577 & 0.593 & 0.508 & 0.459 & 0.413 & -- \\
&AdaAtt & 1.568 & 0.576 & 0.588 & 0.505 & 0.446 & 0.409 & -- \\ 
&Att2in & 1.564 & 0.576 & 0.587 & 0.503 & 0.447 & 0.403 & 25.937 \\
& HRG & 2.800 & 0.588 & 0.629 & 0.547 & 0.497 & 0.463  & -- \\ 
& HRGR-Agent & \textbf{2.895} & 0.612 & \textbf{0.673} & 0.587 & 0.530 & \textbf{0.486} & -  \\ 
\cline{2-9}  
& KER & 0.817 & 0.552 & 0.609 & 0.489 & 0.400 & 0.335 & -- \\ 
& KERP & 2.850 & \textbf{0.618} & \textbf{0.673} & \textbf{0.588} & \textbf{0.532} & 0.473 & \textbf{67.820} \\ 
\hline 
\end{tabular} 
\caption{Automatic and human evaluation on IU X-Ray (upper part) and CX-CHR dataset (lower part) compared with CNN-RNN~\protect\cite{vinyals2015show}, LRCN~\protect\cite{lrcn2015}, AdaAtt~\protect\cite{lu2017knowing}, Att2in~\protect\cite{rennie2016self}, CoAtt~\protect\cite{jing2017automatic}, and HRGR-Agent~\cite{li2018hybrid}. * indicates re-training and evaluation on our data split. }
\label{tab:report-generation} 
\end{table*}

\section{Experiments \& Results} 
%
\subsubsection{Dataset.} 
We conduct experiments on two medical image report datasets. First, \textbf{Indiana University Chest X-Ray Collection (IU X-Ray)}~\cite{iuxray} is a public dataset consisting of 7,470 chest x-ray images paired with their corresponding diagnostic reports. Each patient has 2 images (a frontal view and a lateral view) and a report which includes impression, finding and indication sections. We preprocess the reports by tokenizing and converting to lower-cases. We filter tokens by minimum frequency 3, which results in 1185 unique tokens covering over 99.0\% word occurrences in the corpus. We collect 80 abnormalities and 87 templates for IU X-Ray dataset. 
\textbf{CX-CHR} is a private dataset of chest X-ray images with corresponding Chinese reports collected from a professional medical institution for health checking. The dataset consists of 35,609 patients and 45,598 images. Each patient has one or multiple chest x-ray images in different views (e.g., frontal and lateral), and a corresponding Chinese report. We select patients with no more than 2 images and obtain 33,236 patient samples in total which covers over 93\% of the dataset. We preprocess reports by tokenizing 
and filtering tokens whose frequencies are no less than 3, resulting in 1,479 unique tokens. We collect 155 abnormalities and 362 templates for CX-CHR dataset. More details of the dataset, and collection of abnormalities and templates is in Appendix.

On both dataset, we randomly split the data by patients into training, validation and testing by a ratio of 7:1:2. There is no overlap between patients in different sets. 
To fairly compare with all baselines, we extract visual features for both dataset from a DenseNet~\cite{huang2017densely} jointly pretrained on CX-CHR and public available ChestX-ray8 dataset~\cite{wang2017chestx}. Please see Appendix for details.

\subsubsection{Evaluation metrics.} 
We use area under the curve (AUC) to evaluate performance of abnormality and disease classification.
For evaluating medical report generation, we use two kinds of evaluation metrics: 1) automatic metrics including CIDEr~\cite{cider}, ROUGE-L~\cite{rouge}, BLEU~\cite{bleu}; 2) human evaluation: we randomly select 100 sampled testing results of each method, and conduct surveys through Amazon Mechanical Turk. Each survey question gives a ground truth report, and ask candidate to choose among reports generated by different models. The criteria is the degree of correctness of abnormal findings, language fluency, and content coverage compared to the ground truth report. A default choice is provided in case of no or all reports are preferred. We collect results from 5 participants, and compute the average preference percentage for each model excluding default choices. 

\subsubsection{Training details.} 
We first train \textit{Encode} for abnormality classification, then separately train \textit{Retrieve} with fitted templates supervision, and \textit{Paraphrase} with fitted templates as input and ground truth report as output with fixed \textit{Encode}.
Then we fine-tune \textit{Paraphrase} using sampled templates from \textit{Retrieve}. As \textit{Retrieve} may not predict same length or same order of sentences as ground truth, we re-order the ground truth sentences by choosing the smallest edit distance to the corresponding retrieved template sentence. We use learning rate $1e^{-3}$ for training and $1e^{-5}$ for fine-tuning, and reduce by 10 times when encountering validation performance plateau. We use early stopping, batch size 4 and drop out rate 0.1 for all training.   

Besides, as observed from baseline models which overly predict most popular and normal reports for all testing samples, and the fact that most medical reports describe normal cases, we add post-processing to increase the length and comprehensiveness of the generated reports for both datasets while allowing KERP to focus on abnormal findings. Please refer to Appendix for detailed description.

\begin{table}[t!]
\centering 
\small 
\begin{tabular}{l|l|cc}
\hline  
Dataset & Model  & Abnormality   & Disease     \\ 
\hline
\multirow{3}{*}{\scriptsize{IU X-Ray}}
& DenseNet    		& 0.612   &  0.646      \\
\cline{2-4}  
& Ours-1Graph   & 0.674   &   -    \\
& Ours-2Graphs  & \textbf{0.686}  & \textbf{0.726}  \\ 
\hline 
\hline
\multirow{3}{*}{\scriptsize{CX-CHR}} 
& DenseNet     		& 0.689  & 0.800      \\
\cline{2-4}  
& Ours-1Graph   & 0.721  &   -          \\
& Ours-2Graphs  & \textbf{0.760}  & \textbf{0.862}      \\ 
\hline 
\end{tabular} 
\caption{Classification AUC.}
\label{tab:knowledge-graph-acc}
\end{table}

\begin{figure*}[t] 
\centering 
\includegraphics[scale=0.7]{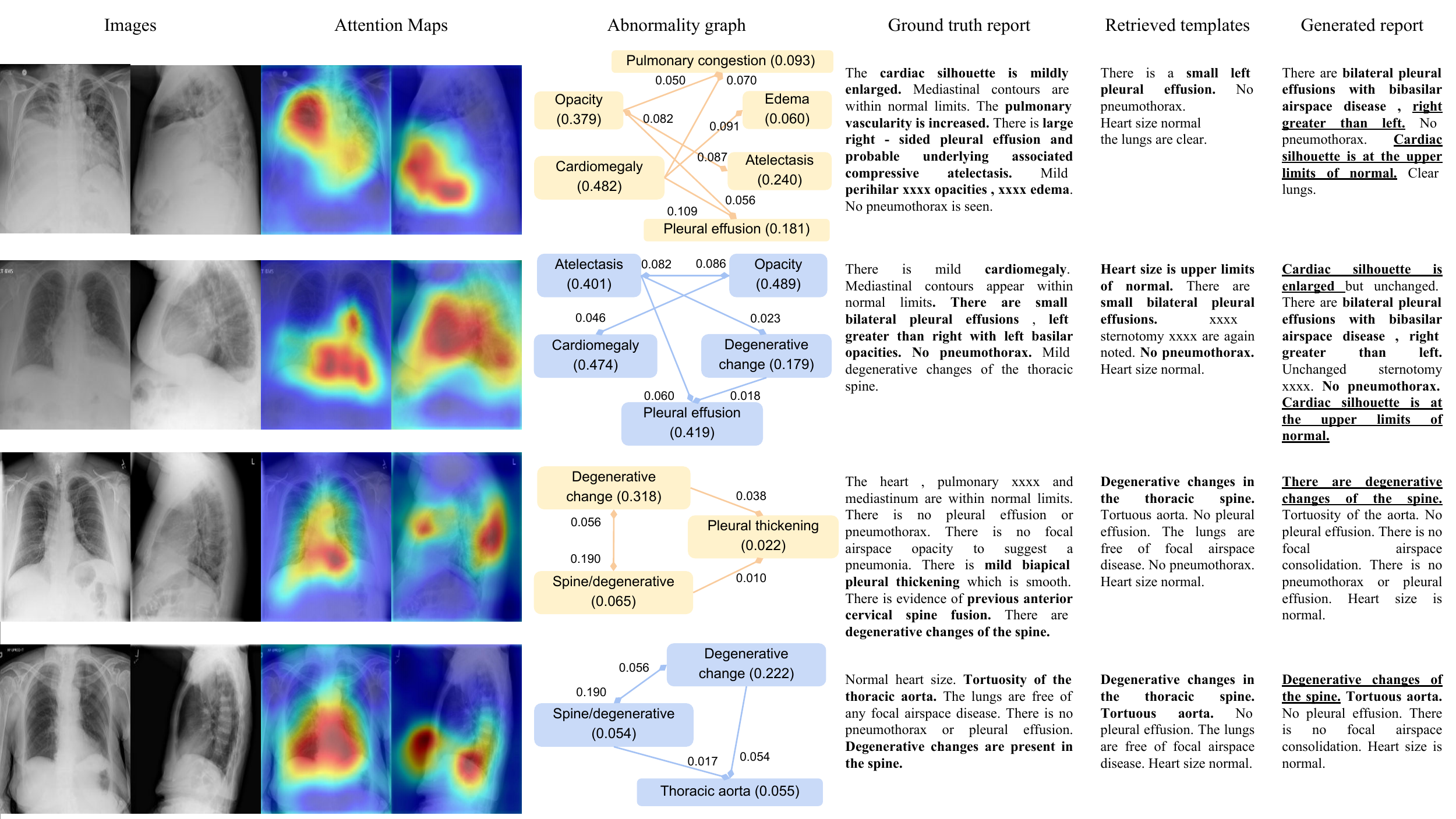} 
\caption{Visualization of result of KERP on IU X-Ray dataset. Bold text indicates alignment between the generated text and ground truth reports. Underlined text indicates correspondence of the generated text (specifically location description) with the visualized attention maps (of either the $1_{st}$/frontal or $2_{nd}$/lateral view). In the abnormality graph, values inside parentheses are predicted probabilities of corresponding nodes. We select edges whose attention scores are greater than or equal to 0.05, or are the highest from neighboring nodes to each node, and visualize the attention scores along the directed arrows. }
\label{fig:qualitative} 
\end{figure*} 
  
\subsubsection{Baselines for abnormality and disease classification} We compare the performance of abnormality and disease classification with a DenseNet~\cite{huang2017densely} trained on classifying the same set of abnormality labels or disease labels. For abnormality classification, we compare our method by solely training on abnormality classification (Ours-1Graph), and jointly training on both abnormality and disease classification (Ours-2Graphs). For disease classification, we directly compare with DenseNet.

\subsubsection{Baselines for medical report generation}
On both datasets, we compare with 4 state-of-the-art image captioning models: CNN-RNN~\cite{vinyals2015show}, LRCN~\cite{lrcn2015}, AdaAtt~\cite{lu2017knowing}, and Att2in~\cite{rennie2016self}. We use the same visual features and train/test split on both datasets respectively. On IU X-Ray dataset, we also compare with CoAtt~\cite{jing2017automatic} which uses different visual features extracted from a pretrained ResNet~\cite{he2016deep}. 
The authors of CoAtt~\cite{jing2017automatic} re-trained their model using our train/test split, and provided evaluation results for automatic report generation metrics using greedy search and sampling temperature 0.5 at test time. 
We also compare with HRGR-Agent~\cite{li2018hybrid} which uses different visual features, same set of train/test split for both dataset, and greedy search and argmax sampling at test time. 
To study the effectiveness of individual modules of KERP, we compare KERP with its variant KER which excludes \textit{Paraphrase} module and only uses retrieved templates as prediction results.

\subsection{Results and Analyses}

\subsubsection{Abnormality and disease classification error analysis.}
The AUCs of abnormality and disease classification are shown in Table~\ref{tab:knowledge-graph-acc}.
Ours-2Graphs outperforms baselines on both dataset on abnormality and disease classification. DenseNet performs the worst on both dataset, and is outperformed by Ours-2Graphs by more than 6\%. This demonstrates that, by incorporating prior medical knowledge and learning internal medical knowledge structure, our method is able to distill useful features for correctly classifying abnormalities and diseases. 
Given the high performance of Ours-2Graphs, we conduct following experiments using the knowledge graph trained with both abnormality and disease labels as initialization.

\subsubsection{Medical report generation error analysis.}  
The results is shown in the upper part of Table~\ref{tab:report-generation}. 
Most importantly, KERP outperforms all baselines on BLEU-1,2,3,4 scores on IU X-Ray dataset, and on ROUGE-L, BLEU-1,2,3 scores on CX-CHR dataset, demonstrating its effectiveness and robustness on combining retrieval and generation mechanism for generating medical reports. 
KERP achieves lower CIDEr score than that of HRGR-Agent on both dataset. However, HRGR-Agent is fine-tuned using reinforcement learning directly with CIDEr as reward. Furthermore, KERP achieves much better performance on abnormality prediction (Table~\ref{tab:knowledge-graph-acc} compared to ~\cite{li2018hybrid}), demonstrating its superior capability of detecting abnormal findings which is important in clinical diagnosis. 
Compared to other baseline models that do not use reinforcement learning, KERP obtains the highest CIDEr score on both dataset. 
On IU X-Ray dataset, KERP achieves lower CIDEr score but higher ROUGE-L and BLEU-n scores than KER which does not have \textit{Paraphrase} process. This indicates that the \textit{Paraphrase} process smooths the retrieved templates into more common sentences as higher BLEU-n scores means higher overlap between n-grams of generated and ground truth reports. However, as CIDEr incorporates inverse document frequency (idf) of words evaluated in the entire evaluation corpus, it inherently gives higher importance to informative and significant phrases, such as abnormal findings and diseases, as oppose to common and popular phrases such as "the lungs are clear" and "heart size is normal". Thus this shows that KER correctly detects more significant phrases, and KERP generates smoother and more common expressions while maintaining the overall performance of abnormal findings detection. 
On CX-CHR dataset, it is observed that KER performs worse than baseline models in most metrics. This may due to the fact that the templates for CX-CHR are designed to be concise and to cover large range of different abnormal findings, instead of being natural and common. Thus only using retrieved templates does not lead to high performance. However, the overall high performance of KERP verifies that \textit{Paraphrase} module is able to distill accurate information from the retrieved templates, and paraphrase them into more common and natural descriptions. It also shows that learning conventional and general writing style of radiologists is as important as accurately detecting abnormalities in medical report generation.

\subsubsection{Human evaluation.} 
We conduct human evaluation as a supplement to automatic evaluation, the result of which is shown in the last column of Table~\ref{tab:report-generation}. KERP outperforms the compared baseline on both dataset, demonstrating its capability of generating fluent and accurate reports. 

\subsubsection{Qualitative analysis.}
The visualization result of KERP on IU X-Ray dataset is shown in Figure~\ref{fig:qualitative}. The generated reports demonstrate significant alignment with ground truth reports as well as correspondence with the visualized attention maps. 
For example, the generated report of the first sample correctly describes "right greater than left" airspace disease, and the attention map of frontal view shows red region over right upper lung greater than left lower lung (the redness indicates degree of attention). The result demonstrates that KERP is capable of generating accurate reports as well as explainable attentive regions. More visualization and analysis on both dataset is in Appendix.

\section{Conclusion}
 
We propose a novel Knowledge-driven Encode, Retrieve, Paraphrase (KERP) method to perform accurate and robust medical report generation, and a generic implementation unit--\textit{Graph Transformer} (GTR) which is the first attempt to transform multi-domain graph-structured data via attention mechanism.  
Experiments show that KERP achieves state-of-the-art performance on two medical image report datasets, and generates accurate attributes prediction, dynamic medical knowledge graph, and explainable location reference. 
 
{
\fontsize{9.0pt}{10.0pt} \selectfont
\bibliography{arxiv}
\bibliographystyle{aaai}
}

\appendixpage 
\section{Dataset statistics}
Detailed statistics of IU X-Ray and CX-CHR dataset is shown in Table~\ref{tab:data-stat}. 
 
\begin{table}[ht]
\centering   \small
\begin{tabular}{l|cc}
\hline  
Statistics                & IU X-Ray  & CX-CHR    \\ \hline
\#Patients                & 3,867     & 35,609     \\
\#Images                  & 7,470     & 45,598     \\ 
\#Diseases                & 14        & 14        \\ 
\#Abnormalities           & 80        & 155       \\  
\#Templates               & 87        & 362       \\ 
\#Abnormal templates      & 66        & 290       \\ 
\#Normal templates        & 21        & 72        \\ 
Vocabulary size             & 1185      & 1479      \\ 
Max. \#sentences      & 18        & 24        \\ 
Max. sentence length & 42   & 38            \\ 
Max. sentence length (by tokens) & 42  & 18              \\ 
Max. report length          & 173       & 216       \\
Avg. \#sentences      &  4.637    & 9.120          \\ 
Avg. sentence length &  6.997  & 7.111            \\ 
Avg. report length          & 32.450    & 64.858          \\ 
\hline 
Avg. predicted \#sentences  & 4.420 & 9.984 \\
Avg. predicted sentence length & 4.751 & 7.309 \\
Avg. predicted report length & 21.003 & 66.045 \\
\hline 
\end{tabular}
\caption{Statistics of CX-CHR and IU X-Ray dataset. "\#" indicates "number of". }
\label{tab:data-stat}
\end{table}

\section{Visual features extraction} 
To fairly compare with all baselines, we extract visual features for both dataset from a DenseNet~\cite{huang2017densely} jointly pretrained on CX-CHR and public available ChestX-ray8 dataset~\cite{wang2017chestx}. IU X-Ray dataset is not used for pretraining due to its relatively small size. 
We add an additional lateral layer as in Feature Pyramid Network~\cite{lin2017feature} for the last three dense blocks and additional convolutional layers to transform feature dimension to 256. We extract features from the last convolutional layer of the second dense block which yields 64 $\times$ 64 $\times$ 256 feature maps. 
Such feature maps contain higher resolution and more location information than that directly extracted from the last convolutional layer of a DenseNet (e.g., 16 $\times$ 16 $\times$ 1024). 

 
\section{Abnormality definition \& collection}
The abnormal findings generally take these forms: 1) the presence of abnormal attributes of an object (e.g., bibasilar consolidation) 
2) absence of typical attributes (e.g., disappearance of costophrenic angle) 
3) abnormal change of object shape (e.g., enlarged heart size) 
4) abnormal change of object location (e.g., elevated left hemidiaphragm). 
We consider all clinical abnormalities stem from thoracic organs as nodes in the abnormality graph. 
We use normalized co-occurrence of abnormality attributes computed from training corpus as prior edge weights for the knowledge graph, and normalized co-occurrence of disease labels as prior edge weights for disease graph, normalized co-occurrence of any abnormality attribute and disease labels as edge weights from source knowledge graph nodes to target disease nodes. The coefficients for prior edge weights mentioned above are all set to 0.9.

\section{Template definition \& collection} 
For each abnormality, we first collect sentences that describe the abnormality and have frequencies no less than a threshold in the training corpus, then manually group sentences of the same meaning and select the most frequent one in each group as template. 
For generating template sequence and report word sequence, we assume no prior edge weights for the target graph or prior edge weights from source graph to target graph. 
We assume uniform prior edge weights among all graph nodes, and use 1.0 weight on the learned edge weights.

\section{Training details} 
$GTR_{i2g}$ for Encode has 3 graph message passing layers and 6 heads in multi-head attention. $GTR_{g2s}$ and $GTR_{gs2s}$ for \textit{Retrieve} and \textit{Paraphrase} respectively has 6 graph message passing layers and 8 heads in multi-head attention. 
The dimension of all hidden features and embedding is set to 256. 
The coefficients for prior edge weights, if provided, are all set to 0.9. 
The word embedding  of \textit{Retrieve} and \textit{Paraphrase}, and the projection matrix $\textbf{W}_w$ to project latent feature to word distribution is shared.
We implement our model by PyTorch and train on two GeForce GTX TITAN GPUs.


\end{document}